%% file: acl2021.tex
\documentclass[11pt,a4paper]{article}
\usepackage[hyperref]{acl2021}
\usepackage{times}
\usepackage{latexsym}

\usepackage{microtype}

\aclfinalcopy 

\usepackage{pifont}

\usepackage{amssymb}
\usepackage{booktabs}
\usepackage{graphicx}
\usepackage{mathtools}
\usepackage{multirow}
\usepackage{paralist}
\usepackage{algorithm}
\usepackage{algpseudocode}
\usepackage{stmaryrd}

\usepackage{xspace}

\newcommand{\ignore}[1]{}

\input{comment_engine.tex}
\newcommand{\DrBoost}{DrBoost\xspace}
\DeclareMathOperator*{\argmax}{arg\,max}

\newcommand{\passage}{\ensuremath{c}}
\newcommand{\question}{\ensuremath{q}}
\newcommand{\corpus}{\ensuremath{\mathcal{C}}}

\newcommand{\dataset}{\ensuremath{\mathcal{D}}}
\newcommand{\datasetWithNegs}{\ensuremath{\widetilde{\mathcal{D}}}}
\newcommand{\trainDataset}{\ensuremath{\dataset_{\text{train}}}}
\newcommand{\devDataset}{\ensuremath{\dataset_{\text{dev}}}}

\newcommand{\trainDatasetWithNegs}{\ensuremath{\datasetWithNegs_{\text{train}}}}
\newcommand{\devDatasetWithNegs}{\ensuremath{\datasetWithNegs_{\text{dev}}}}

\newcommand{\model}{\ensuremath{h}}
\newcommand{\round}{\ensuremath{r}}
\newcommand{\numPassagesInCorpus}{\ensuremath{|\corpus{}|}}
\newcommand{\encoder}{\ensuremath{E}}
\newcommand{\questionEncoder}{\ensuremath{\encoder{}_{Q}}}
\newcommand{\passageEncoder}{\ensuremath{\encoder{}_{C}}}
\newcommand{\questionVec}{\ensuremath{\mathbf{\question}}}
\newcommand{\passageVec}{\ensuremath{\mathbf{\passage}}}

\newcommand*\samethanks[1][\value{footnote}]{\footnotemark[#1]}

\title{Boosted Dense Retriever}

\author{
Patrick Lewis\thanks{\hspace{.06in}Equal contribution}, Barlas Oğuz\samethanks, Wenhan Xiong,  \\
\textbf{Fabio Petroni, Wen-tau Yih, Sebastian Riedel} \\
\\
Meta AI \\
\\
{\tt \{plewis,barlaso,xwhan,fabiopetroni,sriedel,scottyih\}@fb.com} \\
}

\date{}

\begin{document}
\maketitle
\begin{abstract}

We propose \DrBoost, a dense retrieval ensemble inspired by boosting.
\DrBoost is trained in stages: each component model is learned sequentially and \emph{specialized} by focusing only on retrieval mistakes made by the current ensemble. 
The final representation is the concatenation of the output vectors of all the component models, making it a drop-in replacement for standard dense retrievers at test time.
\DrBoost enjoys several advantages compared to standard dense retrieval models. 
It produces representations which are 4x more compact, while delivering comparable retrieval results. It also performs surprisingly well under approximate search with coarse quantization, reducing latency and bandwidth needs by another 4x.  In practice, this can make the difference between serving indices from disk versus from memory, paving the way for much cheaper deployments.

\end{abstract}


\section{Introduction}



Identifying a small number of relevant documents from a large corpus to a given query, information retrieval is not only an important task in-and-of itself, but also plays a vital role in supporting a variety of knowledge-intensive NLP tasks~\cite{lewis_retrieval-augmented_2020, petroni_kilt_2021}, such as open-domain Question Answering~\cite[ODQA,][]{voorhees_building_2000, chen_reading_2017} and Fact Checking~\cite{thorne_fever_2018}.
While traditional retrieval methods, such as TF-IDF and BM25~\cite{robertson_history_2008}, are built on \emph{sparse} representations of queries and documents, \emph{dense} retrieval approaches have shown superior performance recently on a range of retrieval and related large-scale ranking tasks~\cite{guu_retrieval_2020, karpukhin_dense_2020, reimers_sentence-bert_2019, hofstatter_efficiently_2021}.
Dense retrieval involves embedding queries and documents as low-dimensional, continuous vectors, such that query and document embeddings are similar when the document is relevant to the query.
The embedding function leverages the representational power of pretrained language models and is further finetuned using any available training query-document pairs.
Document representations are computed offline in an \emph{index} allowing dense retrieval to scale to millions of documents, with query embeddings being computed on the fly.

When deploying dense retrievers in real-world settings, however, there are two practical concerns: the \emph{size} of the index and the retrieval time \emph{latency}.
The index size is largely determined by the number of documents in the collection, as well as the embedding dimension.
Whilst we cannot generally control the former, reducing the embedding size is an attractive way to reduce index size.
On lowering latency, Approximate Nearest-Neighbor (ANN) or Maximum Inner Product Search (MIPS) techniques are usually required in practice. This implies that it is far more important for retrieval models to perform well under approximate search rather than in the exact search setting.
Developing a dense retrieval model that produces more compact embeddings and are more amenable to approximate search is thus the focus of this research.

In this paper, we propose \DrBoost, an ensemble method for learning a dense retriever, inspired by \emph{boosting}~\cite{schapire_strength_1990,freund_decision-theoretic_1997}. \DrBoost attempts to incrementally build compact representations \emph{at training time}.   
It consists of multiple component dense retrieval models (``weak learners'' in boosting's terminology), where 
each component is a BERT-based bi-encoder, producing vector embeddings of the query and document.
These component embeddings are in lower dimensions (e.g., 32 vs.~768) compared to those of regular BERT encoders.
The final relevance function is implemented as a linear combination of inner products of embeddings produced by each weak learner.
This can be efficiently calculated by concatenating the vectors from each component and then performing a single MIPS search, which makes \DrBoost a drop-in replacement for standard dense retrievers at test time. 
Component models are trained and added to the ensemble \emph{sequentially}. 
Each model is trained as a \emph{reranker} over negative examples sampled by the current ensemble and thus can be seen as specializing on retrieval mistakes made previously. 
For example, early components focus on high-level topical information, whereas later components can capture finer-grained tail phenomena. Through this mechanism, individual components are disentangled and redundancy minimized, leading to more compact representations. 

There are a couple of noticeable differences in training \DrBoost when compared to existing dense retrieval models.
Although iterative training using negatives sampled by models learned in the previous rounds has been proposed~\cite[][inter alia.]{xiong_approximate_2020, qu_rocketqa_2021, oguz_domain-matched_2021, sachan_end--end_2021}, existing methods keep only the final model.
In contrast, the iteratively trained weak learners in \DrBoost are preserved and added to the ensemble.
The construction of the embedding also differs. 
\DrBoost{} can be viewed as a method of slowly ``growing'' overall dense vector representations, lending some structure to otherwise de-localized representations, while existing retrieval models encode queries and documents in one step.

\ignore{
\DrBoost{} is agnostic to the specific dense retrieval algorithm used, and can be applied to any contrastive learning problem with a large set of candidates.
In our experiments, we use DPR~\citep{karpukhin_dense_2020} as the underlying model and investigate web-text and Wikipedia retrieval using MSMARCO~\cite{bajaj_ms_2018} and NaturalQuestions~\cite{kwiatkowski_natural_2019} respectively.
}

More importantly, \DrBoost enjoys several advantages in real-world settings.
Because each weak learner in \DrBoost produces very low-dimensional embeddings to avoid overfitting (32-dim in our experiments),
many components can be added whilst the index stays small.
Our experiments demonstrate that \DrBoost produces very compact embeddings overall, 
achieving accuracy on par with a comparable non-boosting baseline with 4--5x smaller vectors, and strongly outperforming a dimensionally-matched variant.
Probing \DrBoost's embeddings using a novel technique, we also show that the embeddings can be used to recover more topical information from Wikipedia than a dimensionally-matched baseline.

Empirically, \DrBoost performs superbly when using approximate fast MIPS.
With a $k$-mean inverted file index (IVF), the simple and widely used approach, especially in hierarchical indices and web-scale settings \cite{jegou_product_2011, johnson_billion-scale_2019, matsui_paper_2018}, \DrBoost greatly outperforms the baseline DPR model~\cite{karpukhin_dense_2020} by 3--10 points.  
Alternatively, it can reduce bandwidth and latency requirements by 4--64x while retaining accuracy.
In principle, this allows for the approximate index to be served \emph{on-disk} rather than in expensive and limited RAM (which is typically 25x faster), making it feasible to deploy dense retrieval systems more cheaply and at much larger scale.
We also show that \DrBoost's index is amenable to compression, and can be compressed to 800MB, 2.5x smaller than a recent state of the art efficient retriever, whilst being more accurate~\cite{yamada_efficient_2021}.



\section{Dense Retrieval}
\label{sec:dense_retrieval}
Dense Retrieval involves learning a scalable relevance function $\model(\question, \passage)$ which takes high values for passages \passage{} which are relevant for question \question{}, and low otherwise.
In the popular dense bi-encoder framework, $\model(\question, \passage)$ is implemented as the dot product between \questionVec{} and \passageVec{}, dense vector representations of passages and questions respectively, produced by a pair of neural network encoders, \questionEncoder{} and \passageEncoder{},
\begin{equation}
\model(\question,\passage) = \questionEncoder(\question) ^\top \passageEncoder(\passage) = \questionVec ^{\top} \passageVec \label{eq:biencoder}
\end{equation}
where $\questionVec=\questionEncoder(\question)$ and $\passageVec=\passageEncoder(\passage)$.
At inference time, retrieval from a large corpus $\corpus = \{\passage_{_1},\dots,\passage_{_{\numPassagesInCorpus}}\}$ is accomplished by solving the following MIPS problem:
\[
\passage^{*} = \argmax_{\passage \in \corpus}  \questionVec^{\top} \passageVec
\]
In standard settings, we assume access to a set of $m$ gold question-passage pairs~$\dataset = \{(\question_i, \passage_i^+)\}_{i=1}^{m}$.
It is most common to learn models by training to score gold pairs higher than sampled \emph{negatives}.
Negatives can be obtained in a variety of ways, e.g. by sampling at random from corpus \corpus{}, or by using some kind of importance sampling function on retrieval results (see \textsection \ref{sec:iterative_negs}).  
When augmented by $n$ negatives per gold passage-document pair, we have training data of the form:
\[
\datasetWithNegs = \{(\question_i, \passage_i^+, \passage_{i,1}^-,\dots \passage_{i,n}^-)\}_{i=1}^{m}
\]
which we use to train a model, e.g. using a ranking or margin objective, or in our case, by optimizing negative log-likelihood (NLL) of positive pairs 
\[
\mathcal{L}_\theta = -\log \frac{e^{\model(\question_i, \passage_i^+)}}{
e^{\model(\question_i, \passage_i^+)} + 
\sum_{j=1}^n e^{\model(\question_i, \passage_{i,j}^-)}}
\]

\subsection{Iterated Negatives for Dense Retrieval}
\label{sec:iterative_negs}
\begin{algorithm}[t]
\caption{Dense Retrieval with Iteratively-sampled Negatives v.s. Boosted Dense Retrieval}
\label{alg:algorithm}
\small
\begin{algorithmic}[1]
\Require $\trainDataset, \devDataset, \corpus{}$ \Comment{Training Data and Corpus}
\Require $\model_0$ \Comment{Initial Retrieval Model}
\Require  $\tau$ \Comment{Min. Error Reduction Tolerance}
\State $\round \gets 0$ 
\State $\model \gets \model_0$ \Comment{Initialize Current Model with Initial Model}
\State $\epsilon_{\text{old}} \gets \infty$
\State $\epsilon \gets \textbf{GetModelError}(\devDataset, \corpus{}, \model)$
\While{$(\epsilon_{\text{old}} - \epsilon) > \tau$}
\State $\round \gets \round + 1$
\State $\trainDatasetWithNegs^\round = \textbf{AddNegatives}(\trainDataset, \corpus{}, \model)$
\State $\devDatasetWithNegs^\round = \textbf{AddNegatives}(\devDataset, \corpus{}, \model)$
\State $\model_\round = \textbf{TrainDenseRetriever}(\trainDatasetWithNegs^{\round}, \devDatasetWithNegs^{\round})$
\If{Dense Retrieval w/ Iteratively-sampled Negs.}
\State $\model \gets \model_{\round}$
\ElsIf{Boosted Dense Retrieval}
\State $\model \gets \textbf{CombineModels}(\model, \model_{\round})$
\EndIf
\State $\epsilon_{\text{old}} \gets \epsilon$
\State $\epsilon \gets \textbf{GetModelError}(\devDatasetWithNegs, \corpus{}, \model)$

\EndWhile
\State \Return \model
\end{algorithmic}
\end{algorithm}
The choice of negatives is an important factor for what behaviour dense retrievers will learn.
Simply using randomly-sampled negatives has been shown to result in poor outcomes empirically, because they are too easy for the model to discriminate from the gold passage.
Thus, in dense retrieval, it is common to mix in some \emph{hard negatives} along with random negatives, which are designed to be more challenging to distinguish from gold passages~\cite{karpukhin_dense_2020}.
Hard negatives are usually collected by retrieving passages related to a question from an untrained retriever, such as BM25, and filtering out any unintentional golds.
This ensures the hard negatives are at least topically-relevant.

Recently, it has become common practice to run a number of rounds of dense retrieval training to bootstrap hard negatives~\cite[][inter alia.]{xiong_approximate_2020, qu_rocketqa_2021, oguz_domain-matched_2021, sachan_end--end_2021}. 
Here, we first train a dense retriever following the method we describe above, and then use this retriever to produce a new set of hard negatives.
This retriever is discarded, and a new one is trained from scratch, using the new, ``harder'' negatives.
This process can then be repeated until performance ceases to improve.
This approach, which we refer to \emph{dense retrieval with iteratively-sampled negatives} is listed in Algorithm \ref{alg:algorithm}.

\subsection{Boosting}

Boosting is a loose family of training algorithms for machine learning problems, based on the principle of gradually ensembling ``weak learners'' into a strong learner.
Boosting can be described by the following high-level formalism~\cite{schapire_theory_2007}.
For a task with a training set $\{(x_1,y_1),\cdots,(x_m, y_m)\}$, where $(x_i,y_i) \in X \times Y$ we want to learn a function $\model: X\rightarrow Y$, such that $\model(x_i) = \hat{y_i}  \approx y_i$.
This is achieved using an iterative procedure over $R$ steps:
\begin{itemize}
    \item For round $r$, we construct an importance distribution $D_{\round}$ over the training data, based on where error $\epsilon$ of our current model $\model$ is high
    \item Learn a ``weak learner'' $\model_{\round}$ to minimize error $\epsilon_{\round} = \sum_i D_{\round}(i)  \mathcal{L} (\model_{\round}(x_i), y_i)$ for some loss function $\mathcal{L}$ measuring the discrepancy between predictions and real values.
    \item Combine $\model$ and $\model_{\round}$ to form a new, stronger overall model, e.g. by linear combination $\model_{\text{new}} = \alpha \model_\round+ \beta \model $. The iteration can now be repeated.
\end{itemize}
The initial importance distribution $D_0$ is usually assumed to be a uniform distribution, and the $\model_{0}$ model a constant function. 
Note how each additional model added to $\model$ is specifically designed to solve instances that $\model$ currently struggles with.

\subsection{Boosted Dense Retrieval: \DrBoost{}}
\label{sec:boosted_dense_retrieval}
We note similarities between the boosting formulation, and the dense retrieval with iteratively-sampled negatives.
We can adapt a boosting-inspired approach to dense retrieval with minimal changes, as shown in Algorithm~\ref{alg:algorithm}.
Algorithmically, the only difference (lines 10--13) is that in the case of iterative negatives, the model $\model$ after $r$ rounds is \emph{replaced} by the new model $\model_\round$,
whereas in the boosting case, we \emph{combine} $\model_\round$ and $\model$.

In this paper, we view the boosted ``weak learner'' models  $\model_\round$ as \emph{rerankers} over the retrieval distribution from the current model $\model$. 
That is, when training dense boosted retrievers, we \emph{only} train using hard negatives, and do not use any random or in-batch negatives.
Thus each new model is directly trained to solve the retrieval mistakes that the current ensemble makes, strengthening the connection to boosting, using the construction of negatives as a mechanism to define the importance distribution. 

Each model $\model_\round$ is implemented as a bi-encoder, as in Equation~\eqref{eq:biencoder}. We combine models as linear combinations:
\[
\small{\textbf{CombineModels}}(\model, \model_\round) = \model_{\text{new}} = \alpha \model_\round + \beta \model \\
\]
The coefficients could be learnt from development data, or, simply by setting all coefficients to 1, which we find to be empirically effective.
The overall model after the $R$ rounds can be written as:
\begin{align*}
\model(\question, \passage) &= \alpha \model_{_R}(\question, \passage) + \beta \left(\model_{_{R-1}}(\question{}, \passage{})  +  \gamma \left( \cdots \right) \right)\\ 
&=\sum_{\round=1}^{R} \alpha'_{\round} \model_{\round}(\question, \passage) =\sum_{\round=1}^{R} \alpha'_{\round} \questionVec_{\round} ^{\top} \passageVec_{\round}\\
&= [\questionVec_{_R},\dots,\questionVec_{_1},\questionVec_{_0} ]^{\top} [\alpha'_{_R} \passageVec_{_R}, \dots,\alpha'_{_1} \passageVec_{_1},\alpha'_{_0} \passageVec_{_0}]\\
&= \bar{\questionVec}^{\top} \bar{\passageVec}
\end{align*}
where $[\dots]$ indicates vector concatenation. Thus $\model$ is fully decomposable, and can be computed as a single inner product, and as such, is a drop-in replacement for standard MIPS dense retrievers at test time.
The vectors produced by each boosted dense retriever component can be considered to be subvectors of \emph{overall} dense vector representations $\bar{\questionVec}$ and $\bar{\passageVec}$ that are being ``grown'' one round at a time, imparting some structure to the overall vectors.

One downside of the boosting approach is that we must maintain $R$ encoders for both passages and questions. 
Since passages are embedded offline, this does not create additional computational burden on the passage side at test time.
However, on the query side, for a question $\question$, boosted dense retrieval requires $R$ forward passes to compute the full representation, one for each subvector $\questionVec_\round$. This is expensive, and will result in high-latency search.  While this step is fully parallelizable, it is still undesirable.
We can remedy this for low-latency, low-resource settings by distilling the question encoders of $h$ into a single encoder, which can produce the overall question representations $\bar{\questionVec}$ directly.
Here, given the training dataset $\trainDataset$ of gold question-passage pairs, and a model $h$ we want to distill, we first compute overall representations $\bar{\questionVec{}}
$ and $\bar{\passageVec{}}$ for all pairs using $h$ as distillation targets, then train a new question encoder $\questionEncoder^{\text{dist}}$ with parameters $\phi$, by minimizing the objective:
\[
\mathcal{L}_\phi = \sum_{(\question, \passage^+) \in \trainDataset} \lVert \questionEncoder^{\text{dist}}(\question) - \bar{\questionVec} \rVert^2 + \lVert \questionEncoder^{\text{dist}}(\question) - \bar{\passageVec} \rVert^2
\]

\section{Experiments}

\subsection{Datasets}

We train models to perform retrieval for the following two tasks:

\paragraph{Natural Questions (NQ)}\ We evaluate retrieval for downstream ODQA using the widely-used NQ-open retrieval task~\cite{kwiatkowski_natural_2019}.
This requires retrieving Wikipedia passages which contain answers to questions mined from Google search logs.
Gold passages are annotated by crowdworkers as containing answers to given questions.
The retrieval corpus consists of 21M passages, each 100 words in length.
We use the preprocessed and gold pairs prepared by~\citet{karpukhin_dense_2020}, and report recall-at-K (R@K) for K $\in \{20,100\}$.

\paragraph{MSMARCO} We evaluate in a web-text setting using the widely-used passage retrieval task from MSMARCO~\cite{bajaj_ms_2016}.
Queries consist of user search queries from Bing, with human-annotated gold relevant documents.
The corpus consists of 8.8M passages, and we use the preprocessed corpus and training and dev data gold pairs and data splits from \citet{oguz_domain-matched_2021}.
We follow the common practice of reporting the  Mean-Reciprocal-Rank-at-10 (MRR@10) metric for the public development set.

\subsection{Tasks}
In this section, we'll describe the experiments we perform, and the motivations behind them.

\paragraph{Exact Retrieval}\
We are interested in understanding whether the boosting approach results in superior performance for exhaustive (exact) retrieval. 
Here, no quantization or approximations are made to MIPS, which results in large indices, and slow retrieval, but represents the upper bound of accuracy. 
This setting is the most commonly-reported in the literature.

\paragraph{Approximate MIPS: IVF} 
Exact Retrieval does not evaluate how a model performs in practically-relevant settings.
As a result, we also evaluate in two approximate MIPS settings.
First, we consider approximate MIPS with an Inverted File Index~\cite[IVF,][]{sivic_video_2003}.
IVF works by first clustering the document embeddings offline using K-means~\cite{lloyd_least_1982}, 
%
At test time, for a given query vector, rather than compute an inner product for each document in the index, we instead compute inner products to the K centroids. 
We then visit the \texttt{n\_probes} highest scoring clusters, and compute inner products for only the documents in these clusters.
This technique increases the \emph{speed of search} significantly, at the expense of some accuracy.
Increasing $K$, the number of centroids, increases speed, at the expense of accuracy, as does decreasing the value of \texttt{n\_probes}.
A model is preferable if retrieval accuracy remains high with very fast search, i.e. low \texttt{n\_probes} and high $K$ \footnote{Up to the point in $K$ where the first stage search becomes the bottleneck.  This happens when $K$ is in the order of $\sqrt{|\corpus|}$, which is how we pick $K = 65536$.  We also include results with $K \in {4092, 16384}$ in the Appendix.}.
In our experiments we fit $K = 65536$ clusters and sweep over a range of values of \texttt{n\_probes} from  $2^0$ to $2^{15}$.
Other methods such as HNSW~\cite{malkov_efficient_2020} are also available for fast search, but are generally more complex and can increase index sizes significantly.
IVF is a particularly popular approach due it its simplicity, and as a first \emph{coarse} quantizer in hierarchical indexing~\cite{johnson_billion-scale_2019}, since it is straightforward to apply sharding to the clusters, and further search indices can be built for each cluster.

\paragraph{Approximate MIPS: PQ}
Whilst IVF will increase search speeds, it does not reduce the size of the index, which may be important for scalability, latency, and memory bandwidth considerations.
To investigate whether embeddings are amenable to compression, we experiment with applying Product Quantization~\cite[PQ,][]{jegou_product_2011}.
PQ is a lossy quantization method that works by 1) splitting vectors into subvectors 2) clustering each subvector space and 3) representing vectors as a collection cluster assignment codes.
For further details, the reader is referred to \citet{jegou_product_2011}.
We apply PQ using 4-dimensional sub-vectors and 256 clusters per sub-space, leading to a compression factor of 16x over uncompressed \texttt{float32}.

All MIPS retrieval is implemented using FAISS~\cite{johnson_billion-scale_2019}.

\paragraph{Generalization Tests}\ 
In addition to in-domain evaluation, we also perform two generalization tests.
These will determine whether the boosting approach is superior to iteratively-sampling negatives in out-of-distribution settings.
We evaluate MSMARCO-trained models for zero-shot generalization using selected BEIR~\cite{thakur_beir_2021} datasets that have binary relevance labels. Namely, we test on the SciFact, FiQA, Quora and ArguAna subsets.
This will test how well models generalize to new textual domains, and different query surface forms.
We also evaluate NQ-trained models on EntityQuestions~\cite{sciavolino_simple_2021}, a dataset of simple entity-centric questions which has been recently shown to challenge dense retrievers.
This dataset uses the same Wikipedia index as NQ, and tests primarily for robustness and generalization to new entities at test time.

\subsection{Models}

We compare a model trained with iteratively-sampled negatives to an analogous model trained with boosting, which we call \DrBoost{}.
There are many dense retrieval training algorithms available which would be suitable for training  with iteratively-sampled negatives and boosting with \DrBoost{}.
Broadly-speaking, any dense retriever could be used that utilizes negative sampling, and could be trained in step 9 of algorithm \ref{alg:algorithm}.
We choose Dense Passage Retriever~\cite[DPR,][]{karpukhin_dense_2020} with iteratively-sampled negatives due to its comparative simplicity and popularity.

\subsubsection{Iteratively-sampled negatives baseline: DPR} 
DPR follows the dense retrieval paradigm outlined in section \ref{sec:dense_retrieval}
It is trained with a combination of \emph{in-batch} negatives, where gold passages for one question are treated as negatives for other questions in the batch (which efficiently simulates random negatives), and with hard negatives, sampled initially from BM25, and then from the previous round, as in Algorithm \ref{alg:algorithm}.
We broadly follow the DPR training set-up of~\citet{oguz_domain-matched_2021}.
We train BERT-base DPR models using the standard 768 dimensions, as well as models which match the final dimension size of \DrBoost{}.
We use parameter-sharing for the bi-encoders, and layer-norm after linear projection.
Models are trained to minimize the NLL of positives, and the number of training rounds is decided using development data, as in Algorithm \ref{alg:algorithm}, using an initial $h_0$ retriever BM25.

\subsubsection{\DrBoost{} Implementation}
For our \DrBoost{} version of DPR, we keep as many experimental settings the same as possible.
There are two exceptions, which are required for adapting dense retrieval to boosting. 
The first is that each component ``weak learner'' model has a low embedding dimension.
This is to avoid overfitting, ensures each model is not too powerful, and means the final index size is a manageable size.
We report using models of 32-dim (c.f. the standard 768 dim), but note that training with dimension as low as 8 is stable.
The second is that, as motivated in section \ref{sec:boosted_dense_retrieval}, we train each weak learner using \emph{only} hard negatives, and no in-batch negatives. 
In effect, this choice of negatives means each model is essentially trained as a reranker.\footnote{Note: We \emph{sample} negatives from the model's retrieval distribution rather than taking the top-K retrieved negatives. We find this improves results for early rounds}
\DrBoost{} models are fit following algorithm \ref{alg:algorithm}, and we stop adding models when development set performance stops improving.
The initial retriever $\model_0$ for \DrBoost{} is a constant function, and thus the initial negatives for \DrBoost{} are sampled at random from the corpus, unlike DPR, which uses initial hard negatives collected from BM25.

\paragraph{\DrBoost{} $\alpha$ Coefficients}
\DrBoost{} combines weak learners as a linear combination.
We experimented with learning the $\alpha$ coefficients using development data, however this did not significantly improve results over simply setting them all to 1, so for the sake of simplicity and efficiency, we report \DrBoost{} numbers with all $\alpha=1.0$.
Empirically, we find the magnitudes of embeddings for \DrBoost{}'s component models to be similar, and thus their inner products do not drastically differ, so one component does not dominate over others.

\paragraph{\DrBoost{} Distillation}
We experiment with distilling \DrBoost{} ensembles into a single model for latency-sensitive applications using the L2 loss at the end of section~\ref{sec:boosted_dense_retrieval}.
We distill a single BERT-base query encoder, and perform early stopping and model selection using development L2 loss.  

\section{Results}

\input{acl-ijcnlp2021-templates/tables/summary_table}

\subsection{Exact Retrieval}
Exact Retrieval results for MSMARCO and NaturalQuestions are shown in Table \ref{tab:summary} in the ``Exact Search'' Column.
We find that our \DrBoost{} version of DPR reaches peak accuracy after 5 or 6 rounds when using 32-dim weak learners (see section \ref{sec:round_ablations} later), leading to overall test-time index of 160/192-dim.
In terms of Exact Search, \DrBoost{} outperforms the iteratively-sampled negatives DPR baseline on MSMARCO by 2.2\%, and trails it by only 0.3\% on NQ R@100, despite having a total dimension 4-5$\times$ smaller.
It also strongly outperforms a dimensionally-matched DPR, by 3\% on MSMARCO, and 1\% NQ R@100, demonstrating \DrBoost{}'s ability to learn high-quality, compact embeddings.
We also quote recent state-of-the-art results, which generally achieve stronger exact search results~\cite{zhang2021adversarial}. 
Our emphasis, however, is on comparing iteratively-sampled negatives to boosting, and we note that State-of-the-art approaches generally use larger models and more complex training strategies than the ``inner loop'' BERT-base DPR we report here. 
Such strategies could also be incorporated into \DrBoost{} if higher accuracy was desired, as \DrBoost{} is largely-agnostic to the training algorithm used. 

\subsubsection{Number of Rounds}
\label{sec:round_ablations}

\input{acl-ijcnlp2021-templates/tables/number_of_rounds_table}
The performance of DPR and \DrBoost{} on MSMARCO for different numbers of rounds are shown in Table \ref{tab:number_of_rounds}.
We find that all models saturate at about 4 or 5 rounds.  Note \DrBoost{} does not need more iterations to train, even though it doesn't use BM25 negatives for the first round.
On NQ, adding a 6\textsuperscript{th} model slightly improves \DrBoost{}'s precision, at the expense of recall (see Table \ref{tab:summary}).

While iterative training is expensive, we find that subsequent rounds are much cheaper than the first round, with the first round taking $\sim$20K steps in our experiments to converge, with additional \DrBoost{} rounds converging after about 3K steps.

\paragraph{Bagging Dense Retrieval} We also trained a simple ensemble of six 32-dim DPR models for NQ, which we compare to our 6$\times$32-dim component \DrBoost{}.
This experiment investigates whether the improvement over DPR is just a simple ensembling effect, or whether it is due to boosting effects and specialization of concerns.
This DPR ensemble performs poorly, scoring 74.5 R@20 (not shown in tables), 6.8\% below the equivalent \DrBoost{}, confirming that the boosting formulation is important, not simply having several ensembled dense retrievers.

\subsection{Approximate MIPS}
\begin{figure}[t!]
    \centering
    \includegraphics[width=0.9\linewidth]{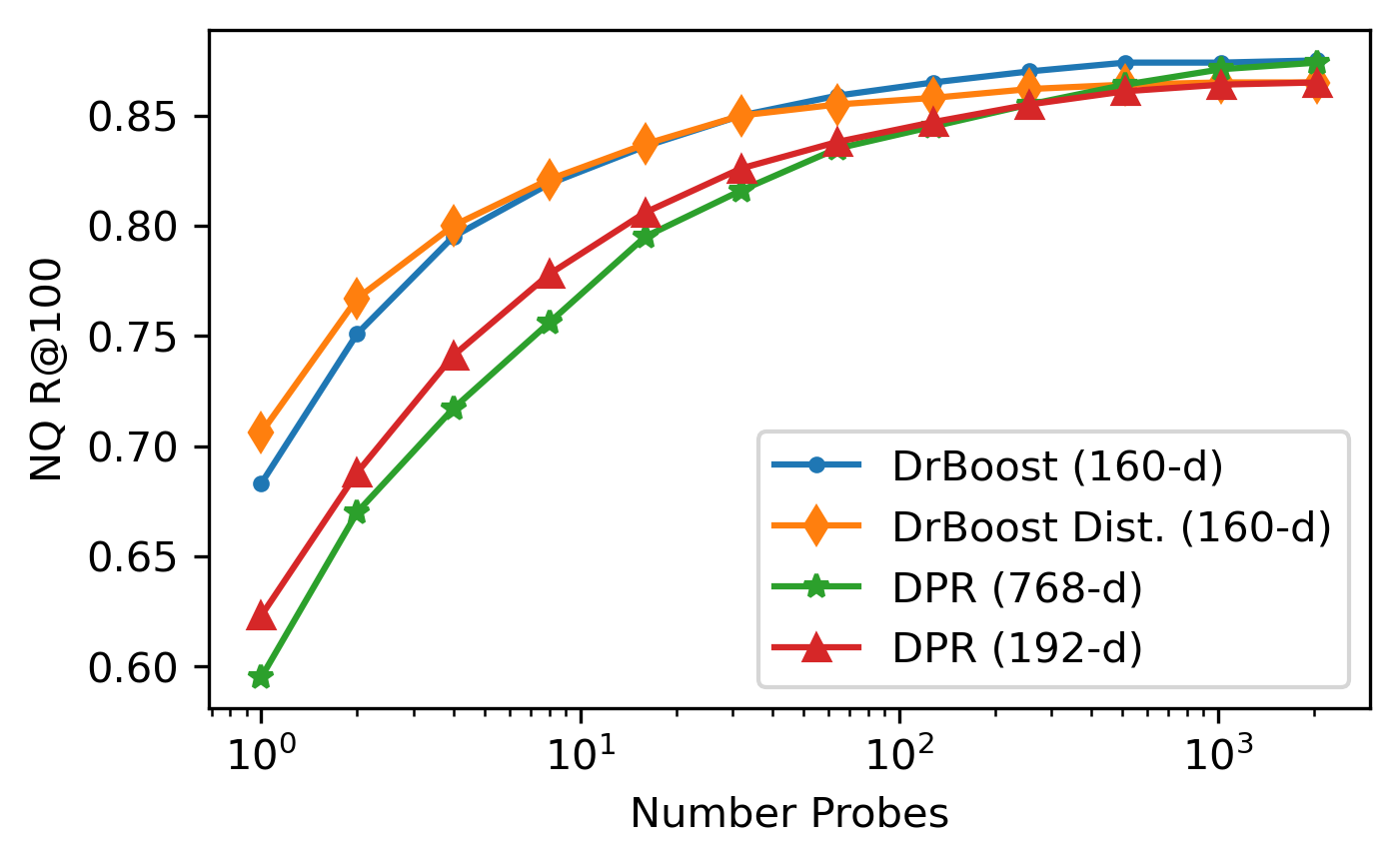}
    \includegraphics[width=0.9\linewidth]{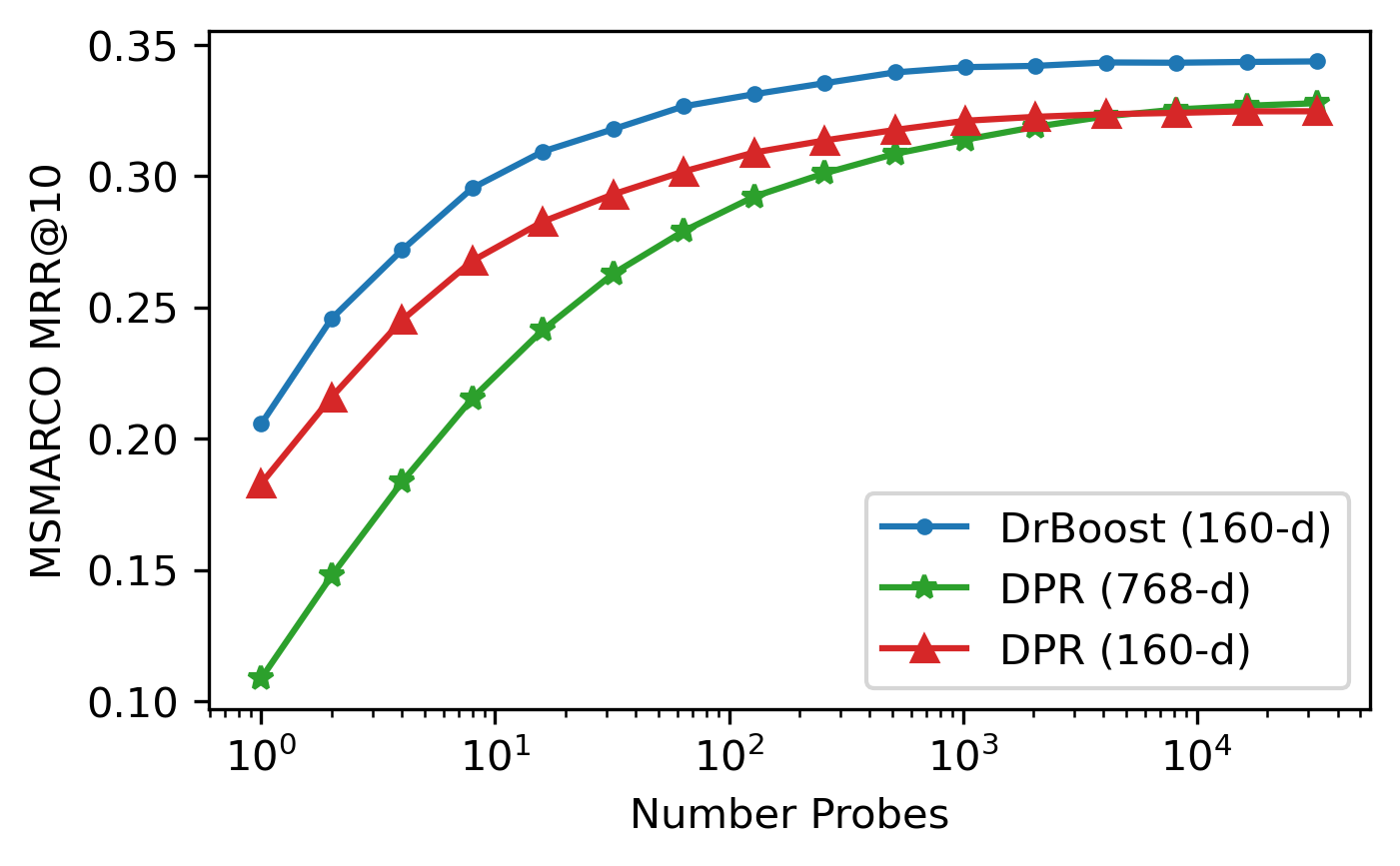}
    
    \caption{Search accuracy vs the number of clusters visited in IVF search (proportional to latency). Accuracy drops as search speed increases, but the accuracy drop-off for \DrBoost{} is much  slower than for DPR.}
    \label{fig:ivf}
\end{figure}
\input{acl-ijcnlp2021-templates/tables/pq_table}

Table \ref{tab:summary} shows how DPR and \DrBoost{} behave under IVF MIPS search, which is also shown graphically in figure \ref{fig:ivf}. 
We find that \DrBoost{} dramatically outperforms DPR in IVF search, indicating that much faster search is possible with \DrBoost{}.
High-dimensional embeddings suffer under IVF due the the curse of dimensionality, thus compact embeddings are important.
Using 8 search probes, \DrBoost{} outperforms DPR by 10.5\% on MSMARCO and 6.3\% on NQ R@100.
The dimensionally-matched DPR is stronger, but still trails \DrBoost{} by about 4\% using 8 probes.
The strongest exact search model is thus not necessarily the best in practical approximate MIPS settings.
For example, if we can tolerate a 10\% relative drop in accuracy from the best performing system's exact search, \DrBoost{} requires 16 (4) probes for MSMARCO (NQ) to reach the required accuracy, whereas DPR will require 1024 (16), meaning \DrBoost{} can be operated approximately 64$\times$ (4$\times$) faster.

The Distilled \DrBoost{} is also shown for NQ in Table \ref{tab:summary}.
The precision (low R@K values) is essentially unaffected, (exact search drops by 0.1\% for R@20), but recall drops slightly (-0.7\% R@100). 
Interestingly, the distilled \DrBoost{} performs even better under IVF search, improving over \DrBoost{} by $\sim$1\% at low numbers of probes.
Crucially, whilst the distilled \DrBoost{} is only slightly better than the 192-dim DPR under exact search, it is 4-5\% stronger under IVF with 8 probes (alternatively, 8$\times$ faster for equivalent accuracy).

Fast retrieval is important, but we may also require small indices for edge devices, or for scalability reasons.
We have already established that \DrBoost{} can produce high quality compact embeddings, but Product Quantization can reduce this even further.
Table \ref{tab:pq_table} shows that \DrBoost{}'s NQ index can be compressed from 13.5 GB to 840MB with less than 1\% drop in performance.
We compare to BPR~\cite{yamada_efficient_2021}, a method specifically designed to learn small indices by learning binary vectors. 
\DrBoost{}'s PQ index is 2.4$\times$ smaller than the BPR index reported by \citet{yamada_efficient_2021}, whilst being 2.4\% more accurate (R@20).
A more aggressive quantization leads to a 420MB index -- 4.8$\times$ smaller than BPR -- whilst only being 1.2\% less accurate.

\section{Analysis}
We conduct qualitative and quantitative analysis to better understand \DrBoost{}'s behavior.

\subsection{Qualitative Analysis}
Since each round's model is learned on the errors of the previous round, we expect each learner to ``specialize" and learn complementary representations.  To see if this is qualitatively true, we look at the retrieved passages from each round's retriever in isolation.  Indeed, we find that each 32-dim sub-vector tackles the query from different angles.  For instance, for the query \textit{``who got the first nobel prize in physics?"}, the first sub-vector captures general topical similarity based on keywords, retrieving passages related to the \textit{``Nobel Prize"}.  The second focuses mostly on the first paragraphs of the Wikipedia articles of prominent historical personalities, presumably because these are highly likely to contain answers in general; and the third one retrieves from the pages of famous scientists and inventors.  The combined \DrBoost{} model would favor passages in the intersection of these sets. Examples can be seen in Table \ref{tab:retrieved-examples} in the Appendix.

\subsection{In-distribution generalization}
\begin{figure}[t!]
    \centering
    \includegraphics[width=0.9\linewidth]{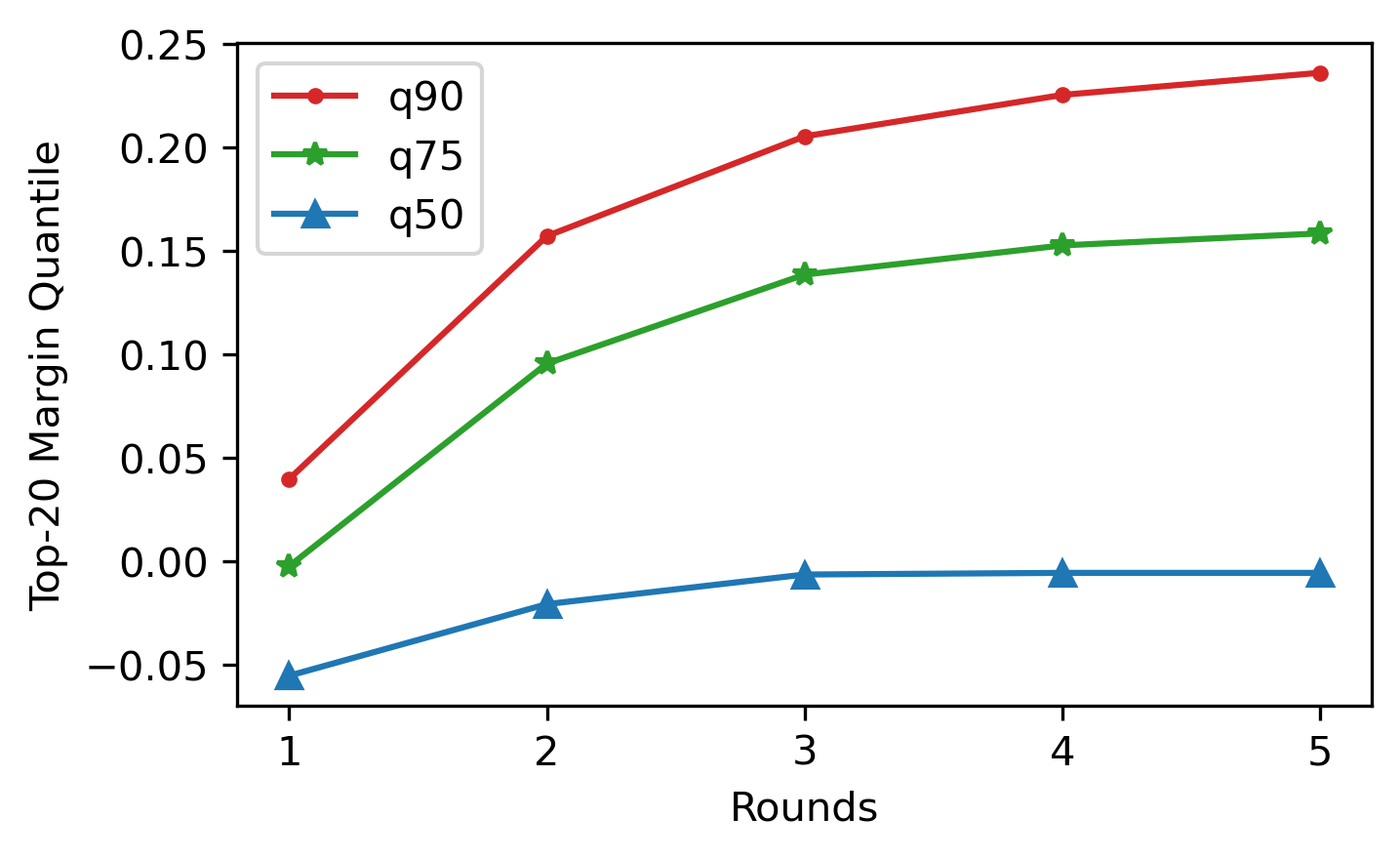}
    \caption{Quantiles of the top-20 margin on the NQ training set, for each iteration of \DrBoost{}.}
    \label{fig:margins}
\end{figure}
Boosting algorithms are remarkably resistant to over-fitting, even when the combined classifier has sufficient capacity to achieve zero training error.  In their landmark paper, \citet{bartlett_boosting_1998} show that this desirable generalization property is a result of the following: the training \emph{margins} increase with each iteration of boosting.  We empirically show the same to be true for \DrBoost{}.  For a given query embedding, dense retrieval acts as a linear classifier, where the gold passage is positive and all other passages are negatives (Eq. \ref{eq:biencoder}).  We adopt the classical definition of margin for linear classifiers to dense retrieval by defining a top-$k$ margin as follows:
\begin{equation}
    \text{Top-k margin}_i = \frac{\model(\question_i,\passage^+) - \max_{c^-}^{\{k\}}\model(\question_i,\passage^-)}{||q_i||\mu_c}
\end{equation}
where $\mu_c$ is the average norm of passage embeddings and the operator $\max^{\{k\}}$ returns the $k$th maximum element in the set.  For a fixed $q_i$ and $k=1$, this definition is identical to the classical margin definition.  Figure \ref{fig:margins} plots the 50th, 75th and 90th percentiles of the top-20 margin for \DrBoost{} on the NQ training set.  We clearly see that margins indeed increase at each step, especially for cases that the model is confident in (high margin).  We hypothesize this property to be the main reason for the strong in-distribution generalization of \DrBoost{} that we observed, and potentially also for the surprisingly strong IVF results, since wide margins should intuitively make clustering easier as well.

\subsection{Cross-domain generalization}
\input{acl-ijcnlp2021-templates/tables/beir_and_ent_questions}
It has been observed in previous work~\cite{thakur_beir_2021} that dense retrievers still largely lag behind sparse retrievers in terms of generalization capabilities. We are interested to test whether our method could be beneficial for out-of-domain transfer as well. We show the results for zero-shot transfer on a subset of the BEIR benchmark in Table~\ref{tab:beir} and the EntityQuestions dataset in Table~\ref{tab:ent_questions}. While \DrBoost{} improves slightly over the dimension-matched baseline on EntityQuestions, where the passage corpora stays the same, it produces worse results on the BEIR datasets. We conclude that boosting is not especially useful for cross-domain transfer, and should be combined with other methods if this is a concern. We leave this for future work.

\subsection{Representation Probing}

\begin{figure}[t!]
    \centering
    \includegraphics[width=\linewidth]{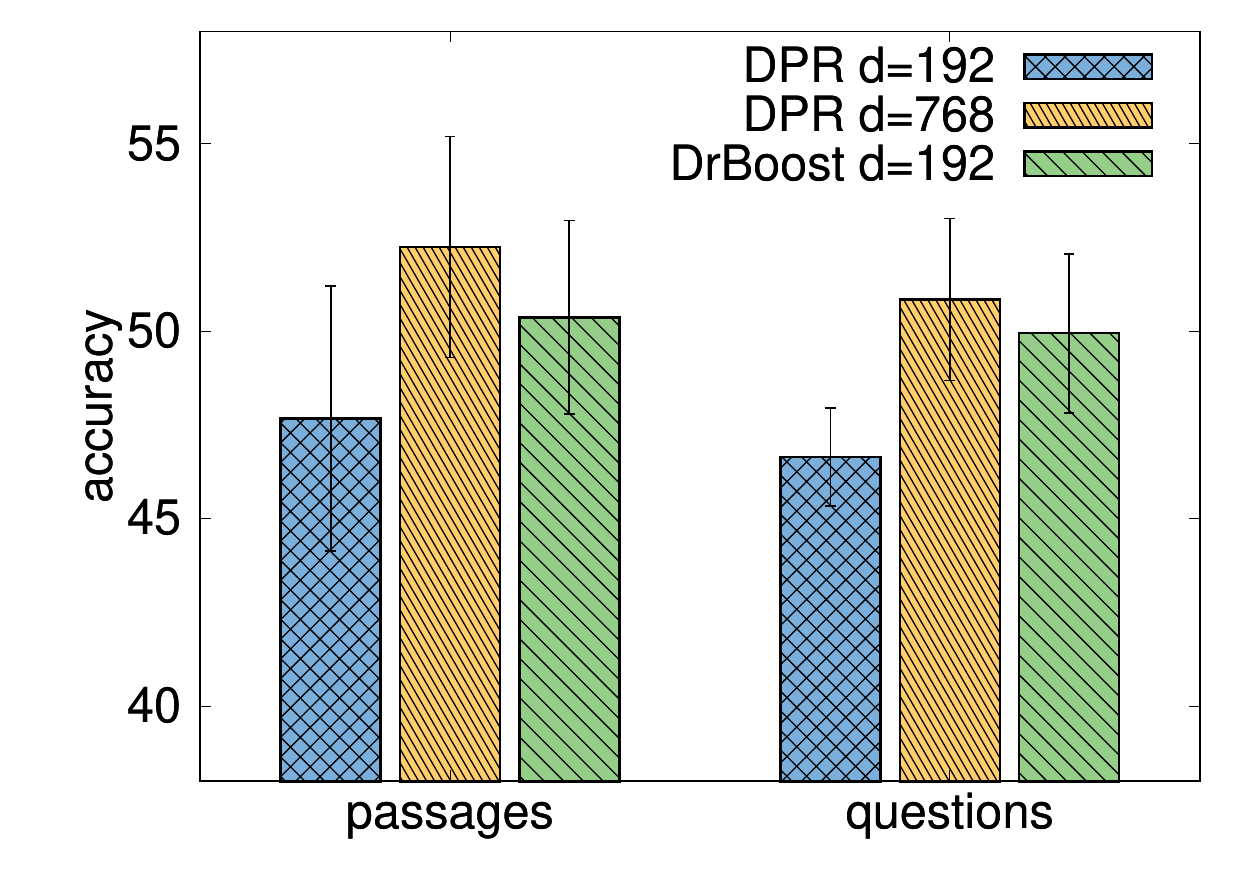}
    \caption{Topic classification accuracy when probing \DrBoost{} and DPR representations with an SVM. }
    \label{fig:topics}
\end{figure}

One of the hypothesis we formulate for the stronger performance of \DrBoost{} over DPR is that the former might better capture topical information of passages and questions. To test this, we collected topics for all Wikipedia articles in Natural Questions using the strategy of \citet{johnson2021language} and associate them with both passages and questions. We then probed both DPR and \DrBoost{} representations with an SVM \cite{steinwart2008support} classifier considering a 5-fold cross-validation over 500 instances and 8 different seeds. Results (in Figure \ref{fig:topics}) confirms our hypothesis: the topic classifier accuracy is higher with \DrBoost{} representations with respect to DPR ones of the same dimension (i.e., 192), for both questions and passages.

\section{Related Work}

\paragraph{Boosting for retrieval} Boosting has been studied in machine learning for over three decades~\cite{kearns_crytographic_1989, schapire_strength_1990}.
Models such as AdaBoost~\cite{freund_decision-theoretic_1997} and GBMs~\cite{friedman_greedy_2001} became popular approaches to classification problems, with implementations such as XGBoost still popular today~\cite{chen_xgboost_2016}.
Many boosting approaches have been proposed for retrieval and learning-to-rank (LTR) problems, typically employing decision trees, such as AdaRank~\cite{xu_adarank_2007}, RankBoost~\cite{freund_efficient_2003} and lamdaMART~\cite{wu_adapting_2009}.
Apart from speed and accuracy, boosting is attractive due to promising theoretical properties such as convergence and generalization.~\cite{bartlett_boosting_1998,freund_efficient_2003,mohri_foundations_2012}. 
Boosted decision trees have recently been demonstrated to be competitive on LTR tasks~\cite{qin_are_2021}, but, in recent years, boosting approaches have generally received less attention, as (pretrained) neural models began to dominate much of the literature. 
However, modern neural models and boosting techniques need not be exclusive, and a small amount of work exploring boosting in the context of modern pre-trained neural models has been carried out~\cite{huang_boostingbertintegrating_2020, qin_are_2021}.
Our work follows this line of thinking, identifying dimensionally-constrained bi-encoders as good candidates as neural weak learners, adopting a simple boosting approach which allows for simple and efficient MIPS at test time.

\paragraph{Dense Retrieval} Sparse, term-based Retrievers such as BM25~\cite{robertson_probabilistic_2009} have dominated retrieval until recently.
Dense, MIPS-based Retrieval using bi-encoder architectures leveraging contrastive training with gold pairs~\cite{yih_learning_2011} has recently shown to be effective in several settings~\cite{lee_latent_2019, karpukhin_dense_2020,reimers_sentence-bert_2019, hofstatter_efficiently_2021}.
See \citet{yates_pretrained_2021} for a survey.
The success of Dense Retrieval has led to  many recent papers proposing schemes to improve dense retriever training by innovating on how negatives are sampled~\cite[][inter alia.]{xiong_approximate_2020, qu_rocketqa_2021, zhan_optimizing_2021, lin_-batch_2021}, and/or proposing pretraining objectives~\cite{oguz_domain-matched_2021,guu_retrieval_2020,chang_pre-training_2020, sachan_end--end_2021, gao_condenser_2021}.
Our work also innovates on how dense retrievers are trained, but is arguably orthogonal to most of these training innovations, since these could still be employed when training each component weak learner.

\paragraph{Distillation} We leverage a simple distillation technique to make \DrBoost{} more efficient at test time. 
Distillation for dense retrievers is an active area, and more complex schemes exist which could improve results further~\cite{izacard_distilling_2021, qu_rocketqa_2021, yang_is_2020, lin_-batch_2021, hofstatter_improving_2021, barkan_scalable_2020, gao_understanding_2020}.

\paragraph{Multi-vector Retrievers}
Several approaches represent passages with multiple vectors. 
\citet{humeau_poly-encoders_2020} represent queries with multiple vectors, but retrieval is comparatively slow as relevance cannot be calculated with a single MIPS call.
ME-BERT~\cite{luan_sparse_2021} index a fixed number of vectors for each passage and ColBERT~\cite{khattab_colbert_2020} index a vector for every word.
Both can perform retrieval with a single MIPS call (although ColBERT requires reranking) but produce very large indices, which, in turn, slows down search.
\DrBoost{} can also be seen as a multi-vector approach, with each weak learner producing a vector.
However, each vector is  small, and we index concatenated vectors, rather than indexing each vector independently, leading to small indices and fast search.
This said, adapting \DrBoost{}-style training to these settings would be feasible.
SPAR~\cite{chen_salient_2021} is a two-vector method: one from a standard dense retriever, and the other from a more lexically-oriented model. 
SPAR uses a similar test-time MIPS retrieval strategy to ours, and SPAR's lexical embeddings could be trivially added to \DrBoost{} as an additional subvector.

\paragraph{Efficient retrievers}
There have been a number of recent efforts to build more efficient retrieval and question answering systems~\cite{min_neurips_2021}.
\citet{izacard_memory_2020} and \citet{yang_designing_2021} experiment with post-hoc compression and lower-dimensional embeddings, \citet{lewis_paq_2021} index and retrieve question-answer pairs and \citet{yamada_efficient_2021} propose BPR, which approximates MIPS using binary vectors.
There is also a line of work learning embeddings specifically suited for approximate search~\cite{yu_product_2018,zhan_jointly_2021,zhan_learning_2021}
Generative retrievers~\cite{de_cao_autoregressive_2021} can also be very efficient.
\DrBoost{} also employs lower-dimensional embeddings and off-the-shelf post-hoc compression for its smallest index, producing smaller indices than BPR, whilst also being more accurate.

\section{Discussion}

In this work we have explored boosting in the context of dense retrieval, inspired by the similarity of iteratively-sampling negatives to boosting.
We find that our simple boosting approach, \DrBoost{}, performs largely on par with a 768-dimensional DPR baseline, but produces more compact vectors, and is more amenable to approximate search.
We note that \DrBoost{} requires maintaining more neural models at test time, which may put a greater demand on GPU resources.
However the models can be run in parallel if latency is a concern, and if needed, these models can be distilled into a single model with little drop in accuracy.
We hope that future work will build on boosting approaches for dense retrieval, including adding adaptive weights, and investigating alternative losses and sampling techniques.
We also suggest that emphasis in dense retrieval should be placed on more holistic evaluation than just exact retrieval accuracy, demonstrating that models with quite similar exact retrieval can perform very differently under practically-important approximate search settings.


\bibliographystyle{acl_natbib}
\bibliography{acl2021}

\appendix
\section{Appendix}

\input{acl-ijcnlp2021-templates/tables/raw_ivf_results}
\input{acl-ijcnlp2021-templates/tables/retrieval_examples}

\end{document}

%% file: comment_engine.tex
\newcommand{\TODO}[1]{}
\newif\ifshowtodo
\showtodotrue
\ifshowtodo
\renewcommand{\TODO}[1]{{\color{red} TODO: {#1}}}
\fi

\newcommand{\CITE}[1]{}
\newif\ifshowtocite
\showtocitetrue
\ifshowtocite
\renewcommand{\CITE}[1]{{\color{blue} CITE: {#1}}}
\fi

\newcounter{srCounter}
\newif\ifsrvar
\srvartrue
\ifsrvar
\newcommand{\seb}[1]{{\small \color{orange} \refstepcounter{srCounter}\textsf{[SR]$_{\arabic{srCounter}}$:{#1}}}}
\else
\newcommand{\seb}[1]{}
\fi

\newcounter{fpCounter}
\newif\iffpvar
\fpvartrue
\iffpvar
\newcommand{\fabio}[1]{{\small \color{blue} \refstepcounter{fpCounter}\textsf{[FP]$_{\arabic{fpCounter}}$:{#1}}}}
\else
\newcommand{\fabio}[1]{}
\fi

\newcounter{plCounter}
\newif\ifplvar
\plvartrue
\ifplvar
\newcommand{\patrick}[1]{{\small \color{magenta} \refstepcounter{plCounter}\textsf{[PL]$_{\arabic{plCounter}}$:{#1}}}}
\else
\newcommand{\patrick}[1]{}
\fi

\newcounter{boCounter}
\newif\ifbovar
\bovartrue
\ifbovar
\newcommand{\barlas}[1]{{\small \color{pink} \refstepcounter{boCounter}\textsf{[BO]$_{\arabic{boCounter}}$:{#1}}}}
\else
\newcommand{\barlas}[1]{}
\fi

\newcounter{wxCounter}
\newif\ifwxvar
\wxvartrue
\ifwxvar
\newcommand{\wenhan}[1]{{\small \color{BlueGreen} \refstepcounter{wxCounter}\textsf{[WX]$_{\arabic{wxCounter}}$:{#1}}}}
\else
\newcommand{\wenhan}[1]{}
\fi

%% file: acl-ijcnlp2021-templates/tables/summary_table.tex
\begin{table*}[t!]
\centering
\resizebox{\textwidth}{!}{   
\begin{tabular}{ccccc|cccc|ccccccccc}
\toprule         
\multirow{4}*{\textbf{Methods}}   & 
\multirow{4}*{\shortstack[c]{\textbf{Total}\\ \textbf{Dimension}}} & 
\multicolumn{3}{c|}{\textbf{MSMARCO}} & 
\multicolumn{8}{c}{\textbf{Natural Questions}} \\
&  & \multicolumn{3}{c|}{\emph{MRR@10}}             & \multicolumn{4}{c|}{\emph{R@20}}        & \multicolumn{4}{c}{\emph{R@100}}         \\
& &  
\multirow{2}*{\shortstack[c]{\small{Exact} \\\small{Search}}} &
\multirow{2}*{\shortstack[c]{\small{IVF} \\\small{8}}} &
\multirow{2}*{\shortstack[c]{\small{IVF} \\\small{64}}} &

\multirow{2}*{\shortstack[c]{\small{Exact} \\\small{Search}}} &
\multirow{2}*{\shortstack[c]{\small{IVF} \\\small{4}}} &
\multirow{2}*{\shortstack[c]{\small{IVF} \\\small{8}}} &
\multirow{2}*{\shortstack[c]{\small{IVF} \\\small{32}}} &

\multirow{2}*{\shortstack[c]{\small{Exact} \\\small{Search}}} &
\multirow{2}*{\shortstack[c]{\small{IVF} \\\small{4}}} &
\multirow{2}*{\shortstack[c]{\small{IVF} \\\small{8}}} &
\multirow{2}*{\shortstack[c]{\small{IVF} \\\small{32}}} &
              \\
&&&&&&&&&&&&\\
\midrule
BM25~\citep{yang_anserini_2017} & - & 18.7 & - & - & 59.1 & - & -   & - & 73.7 & - & - & - \\
AR2~\citep{zhang2021adversarial} & 768  &  39.5  & - & - & 86.0   & - & -   & - &  90.1  & - & - & -\\
\midrule 
\multirow{2}{*}{
DPR w/ iteratively-sampled negatives
} 
 & 768  & 32.8 & 21.6  & 27.9 & \textbf{82.7} & 64.7 & 69.0 & 76.0 & \textbf{87.9} & 71.7 & 75.6 & 81.6\\

 & \multirow{1}*{160 / 192$^*$}
  &  32.5 & 26.8 & 30.2 &  80.8 & 67.9 &  71.7 & 76.6 & 86.6 & 74.1 & 77.8 & 82.6\\
\midrule
\multirow{2}{*}{
\DrBoost{} (32-dim subvectors) 
}
 & 160 (5x32d)  & \textbf{34.4} & \textbf{29.6} & \textbf{32.7} & 80.9 & 73.2 & 75.8 & 78.4 & 87.6 & 79.5 & 81.9 & \textbf{85.0} \\
 & 192 (6x32d) &  -    & -    &  -   & 81.3 & 73.0 & 75.5 & 78.6 & 87.4 & 79.3 & 81.9 & 84.5 \\
\midrule
 \multirow{1}{*}{
\DrBoost{}-Distilled
} & 160 & - & - & - & 80.8 & \textbf{74.4} &  \textbf{76.4} & \textbf{79.3}  &86.8 & \textbf{80.0} & \textbf{82.1} & \textbf{85.0}\\
\bottomrule
\end{tabular}
}
\caption{Summary of Results on MSMARCO development set and NaturalQuestions test set. ``Exact'' indicates Exact MIPS results, IVF indicates IVF MIPS search with 65K centroids, with the number of search probes (proportional to search speed) indicated. $^*$ Dimensional-matched DPR is 160 dims for MSMARCO and 192 for DPR.}
\label{tab:summary}
\end{table*}

%% file: acl-ijcnlp2021-templates/tables/number_of_rounds_table.tex
\begin{table}
\centering
\footnotesize
 \setlength{\tabcolsep}{3pt}

\begin{tabular}{ccccc}
\toprule         
\multirow{2}*{\textbf{Methods}}   & 
\multirow{2}*{\textbf{Round}} & 
\multirow{2}*{\shortstack[c]{\textbf{Total}\\ \textbf{Dim.}}} & 
\multicolumn{1}{c}{\textbf{\textsc{msmarco}}} \\
& & & \multicolumn{1}{c}{\emph{MRR@10}} \\
\midrule
\multirow{6}{*}{
\shortstack[c]{DPR w/ \\ iteratively-sampled negs. \\(Initial Hard Negs. BM25)}
} 
 & 1 & 768  & 28.6         \\
 & 2 & 768  & 32.2   \\
 & 3 & 768  & 32.3              \\
 & 4 & 768  & 32.8   \\
 & 5 & 768  & 32.6   \\
 \cmidrule{2-4}
 & 1 & 160  & 28.9       \\
 & 2 & 160  & 31.4 \\
 & 3 & 160  & 31.7             \\
 & 4 & 160  & 32.1  \\
 & 5 & 160  & 32.3  \\
 & 6 & 160  & 32.5  \\
 & 7 & 160  & 32.3  \\
\midrule
\multirow{5}{*}{
\shortstack[c]{\DrBoost{} \\(32-dim subvectors) \\ (Initial Negs. Random)}
} 
 & 1& 32   & 22.2   \\
 &2& 64   & 31.5   \\
 &3& 96   & 33.8   \\
 &4& 128  & 34.3   \\
 &5& 160  & 34.4  \\
\bottomrule
\end{tabular}
\caption{
Ablations for the number of rounds for DPR with iterative negatives and \DrBoost{} for MSMARCO 
}
\label{tab:number_of_rounds}
\end{table}

%% file: acl-ijcnlp2021-templates/tables/pq_table.tex
\begin{table}
\centering
\footnotesize
 \setlength{\tabcolsep}{2.5pt}
\begin{tabular}{lcccccc}
\toprule     
\multirow{2}*{\textbf{Methods}}   & 
\multirow{2}*{
\shortstack[c]{\textbf{Total}\\ \textbf{Dim.}}
} & 
\multirow{2}*{
\shortstack[c]{\textbf{Size}\\ (GB)}
} & 
 \multicolumn{2}{c}{\textbf{NQ}} \\
& & & \emph{R@20}       & \emph{R@100}         \\
\midrule
DPR \footnotesize{\cite{yamada_efficient_2021}} & 768 & 64.6 & 78.4 & 85.4 \\
\ \ \ + PQ (8-dim subvecs) & &  2.0 & 72.2 & 81.2\\
BPR \footnotesize{\cite{yamada_efficient_2021}} & 768$^*$ & 2.0 & 77.9 & 85.7 \\
\midrule
 \DrBoost{} & 160 & 13.5 & 80.9 & 87.6 \\
\ \ \ + PQ (4-dim subvecs) & & 0.84& 80.3 & 86.8\\
\ \ \ + PQ (8-dim subvecs) & & 0.42 & 76.7 & 84.8\\

\bottomrule
\end{tabular}
\caption{Product Quantization Results. * Indicates Binary vector. 
}
\label{tab:pq_table}
\end{table}

%% file: acl-ijcnlp2021-templates/tables/beir_and_ent_questions.tex
\begin{table}
\centering
\small
\resizebox{0.48\textwidth}{!}{
\begin{tabular}{lcccc}
\toprule     
\multirow{2}*{\textbf{Method}}   & 
 \multicolumn{1}{c}{\textbf{SciFact}} &
  \multicolumn{1}{c}{\textbf{FiQA}} &
 \multicolumn{1}{c}{\textbf{Quora}} & \multicolumn{1}{c}{\textbf{ArguAna}} \\
&  \emph{NDCG@10}&  \emph{NDCG@10} &  \emph{NDCG@10} & \emph{NDCG@10}      \\
\midrule
SotA Dense & 64.3 & 30.8 & 85.2 & 42.9\\
\midrule
DPR (160 dim) & 50.9 & 22.8 & 84.3 & 42.5 \\
\DrBoost{} (160 dim) & 49.7 & 22.4 & 78.8 & 39.9\\
\bottomrule
\end{tabular}
}
\caption{BEIR results. The SotA row is copied from~\citet{thakur_beir_2021}, and the numbers represent the best model for each dataset.}
\label{tab:beir}
\end{table}

\begin{table}
\centering
\footnotesize
\begin{tabular}{lccc}
\toprule     
\multirow{2}*{\textbf{Method}}   & 
 \multicolumn{2}{c}{\textbf{EntityQuestions}} \\
&  \emph{R@20} &  \emph{R@100}       \\
\midrule

BM25~\cite{chen_salient_2021} & 71.2 & 79.7\\
DPR~\cite{chen_salient_2021} & 49.7 & 63.4\\

\midrule
DPR (192 dim) & 47.1 & 60.6\\
\DrBoost{} (160 dim) & 51.2 & 63.4\\
\bottomrule
\end{tabular}

\caption{Entity Questions Results.}
\label{tab:ent_questions}
\end{table}

%% file: acl-ijcnlp2021-templates/tables/raw_ivf_results.tex
\begin{table*}
\centering
\resizebox{\textwidth}{!}{   
\begin{tabular}{cccccccccc}
\toprule
 & \textbf{\DrBoost{}} &  &  & \textbf{DPR} &  &  & \textbf{DPR, 160 dim.} &  & \\
 & (n=4096) & (n=16384) & (n=65536) & (n=4096) & (n=16384) & (n=65536) & (n=4096) & (n=16384) & (n=65536)\\
Exhaustive & 0.3438 & 0.3438 & 0.3438 & 0.328 & 0.328 & 0.328 & 0.3248 & 0.3248 & 0.3248\\
1 & 0.1905 & 0.1884 & 0.2057 & 0.1277 & 0.1186 & 0.1088 & 0.1669 & 0.1637 & 0.183\\
2 & 0.2338 & 0.2359 & 0.2458 & 0.172 & 0.1599 & 0.1479 & 0.2129 & 0.2072 & 0.2159\\
4 & 0.2694 & 0.2652 & 0.2719 & 0.2095 & 0.1996 & 0.1836 & 0.2465 & 0.2395 & 0.2452\\
8 & 0.2919 & 0.2873 & 0.2955 & 0.2433 & 0.2326 & 0.2155 & 0.2722 & 0.2637 & 0.2678\\
16 & 0.3106 & 0.3018 & 0.3094 & 0.2693 & 0.2532 & 0.2415 & 0.2906 & 0.2822 & 0.2827\\
32 & 0.324 & 0.3161 & 0.3179 & 0.2855 & 0.2715 & 0.2629 & 0.3027 & 0.297 & 0.2931\\
64 & 0.3314 & 0.3236 & 0.3266 & 0.2994 & 0.2864 & 0.2791 & 0.3127 & 0.3063 & 0.3018\\
128 & 0.3382 & 0.332 & 0.3312 & 0.31 & 0.2982 & 0.2922 & 0.3179 & 0.3129 & 0.309\\
256 & 0.34 & 0.3375 & 0.3354 & 0.3161 & 0.3092 & 0.3011 & 0.3206 & 0.3182 & 0.3136\\
512 & 0.3424 & 0.34 & 0.3395 & 0.3226 & 0.3141 & 0.3085 & 0.3232 & 0.3212 & 0.3176\\
1024 & 0.3437 & 0.3416 & 0.3415 & 0.325 & 0.3197 & 0.3139 & 0.3243 & 0.3229 & 0.3211\\
2048 & 0.3438 & 0.343 & 0.342 & 0.3279 & 0.3243 & 0.3188 & 0.3247 & 0.3242 & 0.3226\\
4096 &  & 0.3435 & 0.3433 &  & 0.3268 & 0.3228 &  & 0.3249 & 0.3236\\
8192 &  & 0.3438 & 0.3432 &  & 0.3278 & 0.3254 &  & 0.3248 & 0.3241\\
16384 &  &  & 0.3435 &  &  & 0.3268 &  &  & 0.3247\\
32768 &  &  & 0.3437 &  &  & 0.3278 &  &  & 0.3247\\
\bottomrule
\end{tabular}
}
\caption{IVF indexing results on MSMARCO.  Metric is MRR@10. $n$ refers to number of clusters used for IVF training.}
\label{tab:raw_msmarco_ivf_results}
\end{table*}

\begin{table*}
\centering
\resizebox{\textwidth}{!}{   
\begin{tabular}{cccccccc}
\toprule
 & \textbf{\DrBoost{}, 160 dim} &  \textbf{\DrBoost{}-distilled, 160 dim}&  \textbf{\DrBoost{}, 192 dim} &  \textbf{\DrBoost{}-distilled, 192 dim}&  \textbf{DPR} & \textbf{DPR, 192 dim.}  \\
Exhaustive & 0.876 & 0.868 & 0.874 & 0.870 & 0.879 & 0.866  \\
1 & 0.683 & 0.706 & 0.684 & 0.701 & 0.595 & 0.623\\
2 & 0.751 & 0.767 & 0.750 & 0.760 & 0.670 & 0.688\\
4 & 0.795 & 0.800 & 0.793 & 0.803 & 0.717 & 0.741\\
8 & 0.819 & 0.821 & 0.819 & 0.825 & 0.756 & 0.778\\
16 & 0.836 & 0.837 & 0.835 & 0.840 & 0.795 & 0.806\\
32 & 0.850 & 0.849 & 0.845 & 0.848 & 0.816 & 0.826\\
64 & 0.859 & 0.855 & 0.858 & 0.856 & 0.835 & 0.838\\
128 & 0.865 & 0.858 & 0.864 & 0.859 & 0.845 & 0.847\\
256 & 0.870 & 0.862 & 0.868 & 0.863 & 0.855 & 0.855\\
512 & 0.874 & 0.864 & 0.870 & 0.866 & 0.864 & 0.861\\
1024 & 0.874 & 0.865 & 0.871 & 0.866 & 0.871 & 0.864\\
2048 & 0.875 & 0.865 & 0.873 & 0.867 & 0.874 & 0.865\\
\bottomrule
\end{tabular}
}
\caption{IVF indexing results on NQ.  Metric is Recall@100. The number of clusters used for IVF training was 65536.}
\label{tab:raw_nq_ivf_results_100}
\end{table*}
\begin{table*}
\centering
\resizebox{\textwidth}{!}{   
\begin{tabular}{cccccccc}
\toprule
 & \textbf{\DrBoost{}, 160 dim} &  \textbf{\DrBoost{}-distilled, 160 dim}&  \textbf{\DrBoost{}, 192 dim} &  \textbf{\DrBoost{}-distilled, 192 dim}&  \textbf{DPR} & \textbf{DPR, 192 dim.}  \\
Exhaustive & 0.809 & 0.809 & 0.813 & 0.809 & 0.827 & 0.808 \\
1 & 0.624 & 0.650 & 0.625 & 0.647 & 0.518 & 0.557\\
2 & 0.686 & 0.703 & 0.684 & 0.703 & 0.597 & 0.625\\
4 & 0.732 & 0.744 & 0.730 & 0.746 & 0.647 & 0.679\\
8 & 0.758 & 0.764 & 0.755 & 0.764 & 0.690 & 0.717\\
16 & 0.771 & 0.779 & 0.775 & 0.780 & 0.732 & 0.743\\
32 & 0.784 & 0.793 & 0.786 & 0.791 & 0.760 & 0.766\\
64 & 0.794 & 0.797 & 0.799 & 0.799 & 0.779 & 0.780\\
128 & 0.799 & 0.800 & 0.805 & 0.801 & 0.791 & 0.789\\
256 & 0.804 & 0.803 & 0.810 & 0.804 & 0.803 & 0.797\\
512 & 0.807 & 0.805 & 0.812 & 0.807 & 0.813 & 0.803\\
1024 & 0.808 & 0.806 & 0.812 & 0.807 & 0.820 & 0.805\\
2048 & 0.808 & 0.806 & 0.813 & 0.808 & 0.823 & 0.807\\
\bottomrule
\end{tabular}
}
\caption{IVF indexing results on NQ.  Metric is Recall@20. The number of clusters used for IVF training was 65536.}
\label{tab:raw_nq_ivf_results_20}
\end{table*}
\begin{table*}
\centering
\resizebox{\textwidth}{!}{   
\begin{tabular}{cccccccc}
\toprule
 & \textbf{\DrBoost{}, 160 dim} &  \textbf{\DrBoost{}-distilled, 160 dim}&  \textbf{\DrBoost{}, 192 dim} &  \textbf{\DrBoost{}-distilled, 192 dim}&  \textbf{DPR} & \textbf{DPR, 192 dim.}  \\
Exhaustive &   0.710 & 0.706 & 0.715 & 0.703 & 0.731 & 0.710\\
1 & 0.544 & 0.560 & 0.535 & 0.557 & 0.439 & 0.475\\
2 & 0.597 & 0.615 & 0.594 & 0.605 & 0.506 & 0.540\\
4 & 0.634 & 0.646 & 0.636 & 0.644 & 0.551 & 0.593\\
8 & 0.662 & 0.663 & 0.663 & 0.665 & 0.597 & 0.623\\
16 & 0.678 & 0.676 & 0.681 & 0.680 & 0.640 & 0.653\\
32 & 0.691 & 0.689 & 0.692 & 0.688 & 0.666 & 0.671\\
64 & 0.699 & 0.692 & 0.702 & 0.695 & 0.687 & 0.685\\
128 & 0.704 & 0.696 & 0.708 & 0.699 & 0.698 & 0.694\\
256 & 0.707 & 0.701 & 0.710 & 0.702 & 0.709 & 0.703\\
512 & 0.709 & 0.703 & 0.712 & 0.703 & 0.718 & 0.707\\
1024 & 0.710 & 0.704 & 0.713 & 0.703 & 0.726 & 0.708\\
2048 & 0.710 & 0.704 & 0.714 & 0.704 & 0.728 & 0.709\\
\bottomrule
\end{tabular}
}
\caption{IVF indexing results on NQ.  Metric is Recall@5. The number of clusters used for IVF training was 65536.}
\label{tab:raw_nq_ivf_results_5}
\end{table*}

%% file: acl-ijcnlp2021-templates/tables/retrieval_examples.tex
\begin{table*}[!t]
\scriptsize
\begin{tabular}{p{1cm}|p{7cm}|p{7cm}}
\toprule
\textbf{Rounds} & \textbf{who got the first nobel prize in physics?} & \textbf{when is the next deadpool movie being released?} \\
\midrule
1
& 
\emph{0: Title: Nobel Prize in Physics} \newline
The Nobel Prize in Physics is a yearly award given by the Royal Swedish Academy of Sciences for those who have made the \ldots
\newline
\emph{1: Title: Nobel Prize in Physics} \newline
\ldots receive a diploma, a medal and a document confirming the prize amount. Nobel Prize in Physics \ldots
\newline
\emph{2: Title: Nobel Prize controversies} \newline
\ldots research CERN, commented in a scientific meet in Kolkata titled "Frontiers of Science" that "it is unfortunate that pioneering \ldots
&
\emph{0: Title: Deadpool (film)} \newline
\ldots was written by Reese and Wernick and played in front of "Logan". "Deadpool 2" was released on May 18, 2018, with \ldots
\newline
\emph{1: Title: Deadpool 2} \newline
\ldots chimichangas, traditionally Deadpool's favorite food, as well as "Deadpool"-inspired Harder drinks. The campaign also \ldots
\newline
\emph{2: Title: Deadpool 2} \newline
\ldots the final two hours. By May 2018, Leitch was working on an official extended edition of the film with Fox wanting to "spin that \ldots
 \\
\midrule
2
& 
\emph{0: Title: George B. McClellan} \newline
George Brinton McClellan (December 3, 1826-October 29, 1885) was an American soldier, civil engineer, railroad executive \ldots
\newline
\emph{1: Title: Johannes Brahms} \newline
Johannes Brahms (; 7 May 1833 – 3 April 1897) was a German composer and pianist of the Romantic period. Born in Hamburg \ldots 
\newline
\emph{2: Title: Bede} \newline
Bede ( ; ; 672/3 – 26 May 735), also known as Saint Bede, Venerable Bede, and Bede the Venerable (), was an English Benedictine \ldots 
&
\emph{0: Title: Here and Now (2018 TV series)} \newline
Here and Now is an American drama television series created by Alan Ball. The series consists of ten episodes and \ldots
\newline
\emph{1: Title: Deadpool 2} \newline
\ldots is dedicated to her memory. The film's score is the first to receive a parental advisory warning for explicit content, and \ldots 
\newline
\emph{2: Title: I'm New Here} \newline
I'm New Here is the 13th and final studio album by American vocalist and pianist Gil Scott-Heron. It was released on February \ldots
\\
\midrule
3
& 
\emph{0: Title: Henri Poincare} \newline
Jules Henri Poincaré (; ; 29 April 1854 – 17 July 1912) was a French mathematician, theoretical physicist, engineer, and \ldots \newline
\emph{1: Title: Marie Curie} \newline
\ldots named in her honor. Marie Curie Marie Skłodowska Curie (; ; ; born Maria Salomea Skłodowska; 7 November 18674 July 1934 \ldots
\newline
\emph{2: Title: Alberto Santos-Dumont} \newline
Alberto Santos-Dumont (; 20 July 187323 July 1932, usually referred to as simply Santos-Dumont) was a Brazilian inventor \ldots
&
\emph{0: Title: Deadpool 2} \newline
\ldots is dedicated to her memory. The film's score is the first to receive a parental advisory warning for explicit content, and \ldots
\newline
\emph{1: Title: Deadpool (film)} \newline
\ldots was written by Reese and Wernick and played in front of "Logan". "Deadpool 2" was released on May 18, 2018, with \ldots 
\newline
\emph{2: Title: Kong: Skull Island} \newline
\ldots later moved to Warner Bros. in order to develop a shared cinematic universe featuring Godzilla and King Kong. \ldots
\\
\bottomrule
\end{tabular}
\caption{Example retrieval results from each round of \DrBoost{}.  Only the beginning of each passage is shown.}
\label{tab:retrieved-examples}
\end{table*}